%% file: PaperForCameraReady.tex
\documentclass[10pt,twocolumn,letterpaper]{article}

\usepackage{wacv}              

\usepackage{graphicx}
\usepackage{amsmath}
\usepackage{amssymb}
\usepackage{booktabs}
\usepackage{algorithm}
\usepackage{algorithmic}
\usepackage[switch]{lineno}
\usepackage{multirow}
\usepackage{colortbl}
\usepackage[accsupp]{axessibility}
\usepackage[pagebackref,breaklinks,colorlinks]{hyperref}

\usepackage[capitalize]{cleveref}
\crefname{section}{Sec.}{Secs.}
\Crefname{section}{Section}{Sections}
\Crefname{table}{Table}{Tables}
\crefname{table}{Tab.}{Tabs.}

\def\wacvPaperID{382} 
\def\confName{WACV}
\def\confYear{2024}
\makeatletter
\renewcommand{\paragraph}{%
  \@startsection{paragraph}{4}%
  {\z@}{2ex \@plus 1ex \@minus .2ex}{-1em}%
  {\normalfont\normalsize\bfseries}%
}
\makeatother

\begin{document}

\title{MIDAS: Mixing Ambiguous Data with Soft Labels \\
for Dynamic Facial Expression Recognition}

\author{Ryosuke Kawamura$^1$, Hideaki Hayashi$^2$, Noriko Takemura$^3$, Hajime Nagahara$^2$\\
$^1$Fujitsu Research of America, Inc.\\
$^2$Osaka University \\
$^3$Kyushu Institute of Technology\\
{\tt\small rkawamura@fujitsu.com,  \{hayashi, nagahara\}@ids.osaka-u.ac.jp, takemura@ai.kyutech.ac.jp}}

\maketitle

\begin{abstract}
   Dynamic facial expression recognition (DFER) is an important task in the field of computer vision. To apply automatic DFER in practice, it is necessary to accurately recognize ambiguous facial expressions, which often appear in data in the wild. In this paper, we propose MIDAS, a data augmentation method for DFER, which augments ambiguous facial expression data with soft labels consisting of probabilities for multiple emotion classes. In MIDAS, the training data are augmented by convexly combining pairs of video frames and their corresponding emotion class labels, which can also be regarded as an extension of mixup to soft-labeled video data. This simple extension is remarkably effective in DFER with ambiguous facial expression data. To evaluate MIDAS, we conducted experiments on the DFEW dataset. The results demonstrate that the model trained on the data augmented by MIDAS outperforms the existing state-of-the-art method trained on the original dataset.
\end{abstract}


\section{Introduction}
\input{src/1_Introduction.tex}

\section{Related Work}
\input{src/2_RelatedWork.tex}

\section{MIDAS: Mixing Ambiguous Data with Soft Labels}
\input{src/3_Method.tex}

\section{Experiments}
\input{src/4_Experiments.tex}

\vspace{-2mm}
\section{Analysis and Ablation Studies}
\input{src/5_AblationStudy.tex}

\section{Conclusion}
\input{src/6_Conclusion.tex}

{\small
\bibliographystyle{ieee_fullname}
\bibliography{refs}
}

\end{document}

%% file: src/1_Introduction.tex
\label{sec:intro}
Facial expressions play an important role in human communication, and facial expression recognition (FER) has broad applications in areas such as human-computer interaction, driver monitoring, and intelligent tutoring systems for education. To correctly understand emotions from facial expressions, the temporal cues of facial expressions are important for FER because facial expressions are based on facial muscle movements, as demonstrated in previous research~\cite{liu2014learning,yang2008facial,yang2009boosting}. Accordingly, our study focuses on dynamic FER (DFER), which is the task of recognizing an emotion class from a video clip.

\begin{figure}[t]
  \centering
  \includegraphics[width=0.7\linewidth]{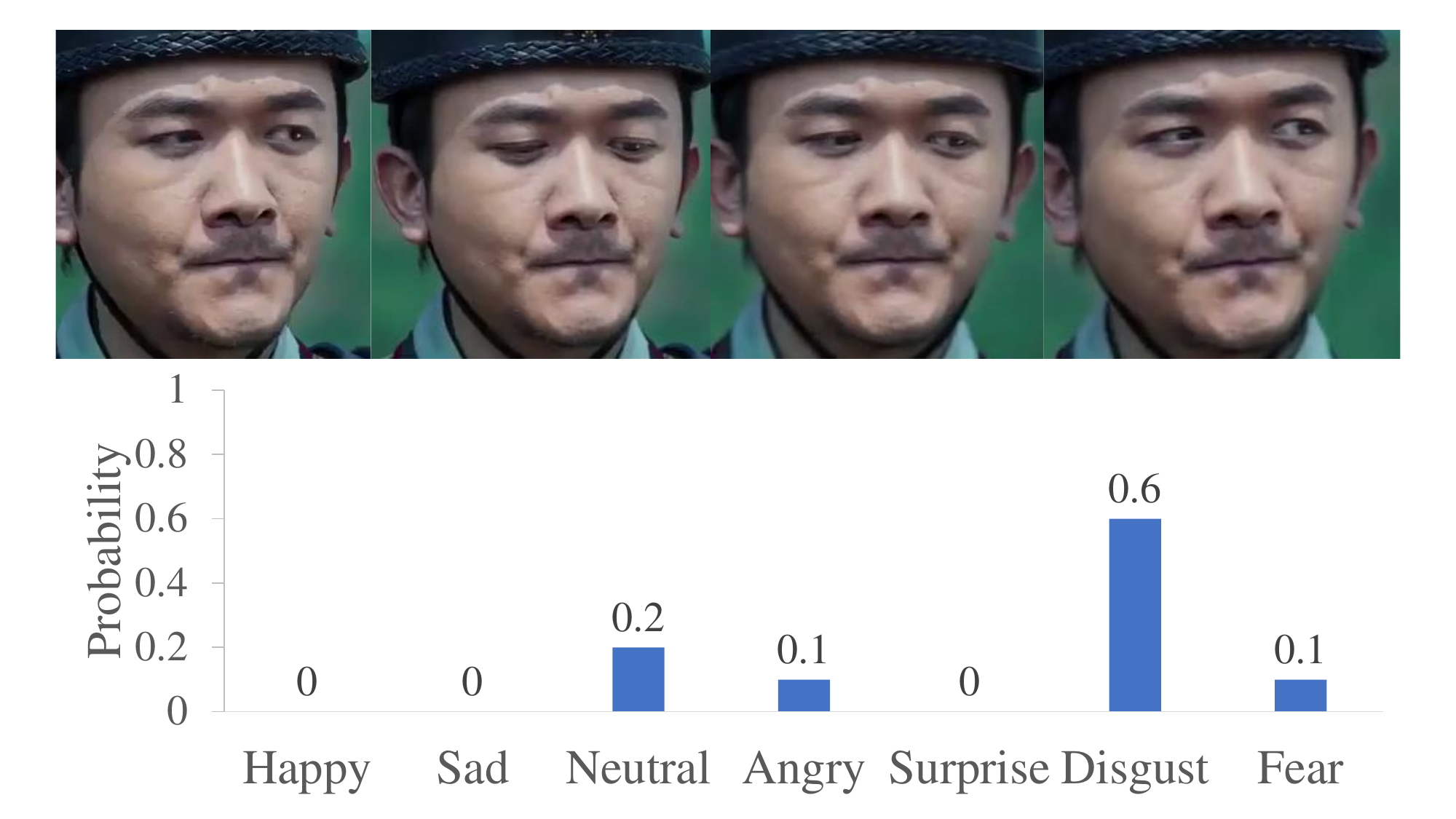}
  \vspace{-3pt}
  \caption{Example of an ambiguous facial expression. The images were taken from the DFEW dataset~\protect\cite{jiang2020dfew}. The bar chart in the bottom row shows the soft-labeled annotation constructed based on the proportions of votes by ten annotators. The annotations are split into four emotion classes.}
  \vspace{-10pt}
  \label{fig:ex}
\end{figure}

Although deep learning-based techniques have shown remarkable performance in DFER on lab-controlled data, DFER on in-the-wild data is still a difficult problem because such data include ambiguous facial expressions, which cannot simply be categorized into a single emotion class. 
There are several factors that contribute to the ambiguity in facial expressions, with the coexistence of multiple emotions being one of the significant factors. 
Since emotions are not mutually exclusive but collective, multiple emotions can coexist at different intensities in ambiguous facial expressions captured under natural conditions. 
This is a significant difference from lab-controlled data, where the researcher usually instructs the subject to make facial expressions. 
In addition, facial expression varies over time, and multiple emotions can be contained even within a single video clip. 
For example, in Fig.~\ref{fig:ex}, the annotators' evaluations for this video clip were split between ``disgust,'' ``neutral,'' and ``fear'' with different probabilities. These features of emotions and facial expressions are considered to be the main factor of ambiguity.

Attaching soft labels to training data, instead of hard labels, is an effective way to address the ambiguity in DFER. Hard labels, that is, one-hot encoded class labels, are mostly used in general recognition tasks such as object recognition, where the input sample is clearly categorized into a single class; however, they cannot appropriately represent an objective variable composed of a combination of multiple emotions with different intensities in ambiguous facial expressions. To correctly learn the ambiguity in DFER, soft labels consisting of probabilities for multiple emotions are helpful to maximize the use of information provided by annotators. One possible method of assigning soft labels to training data is to have multiple annotators evaluate the training data and use the ratio of their votes.

The disadvantage of soft labels is that they are more flexible than hard labels, making it difficult to collect a variety of labels in a uniform manner. There is an enormous amount of possible combinations of emotion classes and the corresponding probabilities, and therefore it is difficult to prepare training data that include all of these patterns. Furthermore, the size of the dataset itself also tends to be limited in DFER due to the difficulty of manual annotation and data collection. To address this problem, it is necessary to augment data effectively and properly to learn from limited data. 


In this paper, we propose a data augmentation method for DFER with ambiguous facial expressions called MIDAS (Mixing Ambiguous Data with Soft labels). In MIDAS, the mixing strategy is expanded to handle soft labels and dynamic facial expressions. The method convexly combines pairs of video frames of facial expressions and the corresponding soft labels that represent the probabilities of emotion classes after aligning the facial position. It then trains a model on the generated data.

Our contributions are summarized as follows:
\begin{itemize}
    \item We proposed MIDAS, a data augmentation method for DFER with ambiguous facial expressions. MIDAS convexly combines pairs of video frames of facial expressions in a similar way to mixup. One significant difference from mixup is that MIDAS is applicable when the true hard labels are unknown and only soft labels consisting of multiple classes of probabilities are given. 
    \item We showed that MIDAS corresponds to minimizing the vicinal risk in a situation where the true hard label is unknown with a vicinity distribution using a random ratio and virtual labels that are different from the original mixup.
    \item Through DFER experiments on the DFEW dataset, we showed that the proposed method outperforms existing state-of-the-art methods. Through an ablation study, we also showed that the combination of soft labels and mixing strategy has a synergistic effect although the use of each individually is also effective. Additionally, the effectiveness of MIDAS with hard labels was demonstrated on both the DFEW and FER39k datasets.
\end{itemize}

%% file: src/2_RelatedWork.tex
\subsection{DFER}\label{sec:rw-dfer}
While many in-the-wild datasets for static FER utilize images collected from the internet, most datasets for DFER, including CK+~\cite{luceyExtendedCohnKanadeDataset2010} and Oulu-CASIA~\cite{zhao2011facial}, are created under lab-controlled environments. In these datasets, the changes in facial expression are prompted by researchers' instructions. Although still relatively few, there has been a growing trend toward the development of large-scale in-the-wild datasets for DFER. Notably, AFEW, introduced by Dhall \textit{et al.}~\cite{dhallCollectingLargeRichly2012} stands out as the pioneering in-the-wild DFER dataset, comprising short clips from movies annotated by a pair of annotators. Similarly, Jiang \textit{et al.}~\cite{jiang2020dfew} collected movie clips for the DFEW datasets. They assigned ten out of twelve annotators to one video clip. This dataset uniquely provides both single-labeled and seven-dimensional emotion class annotations. Furthermore, FERV39k~\cite{wang2022ferv39k} presents a large in-the-wild dataset tailored for DFER. This dataset contains video clips in 22 fine-grained contexts such as business, daily life, and school. The data are annotated by 20 crowd-sourcing annotators and 10 professional researchers.

Regarding FER in the video, methods based on selecting peak frames or aggregating features from each frame have been proposed by~\cite{zhao2016peak,meng2019frame,knyazev2017convolutional,yu2018deeper,yang2018facial,kumar2020noisy}. Two- or three-dimensional convolutional neural networks (2D-CNN or 3D-CNN) combined with a sequential neural network, such as long short-term memory (LSTM) and gated recurrent unit (GRU), are commonly used in~\cite{kim2017multi,yan2018multi,chen2020stcam,kim2017deep,vielzeuf2017temporal,zhang2020facial,liu2020saanet,aminbeidokhti2019emotion}. 

Several studies have proposed the utilization of a Transformer-based module for DFER~\cite{zhao2021former,ma2022spatio}. For example, Zhao \textit{et al.}~\cite{zhao2021former} proposed Former-DFER, which is based on the Transformer module, and Ma \textit{et al.}~\cite{ma2022spatio} used features processed by a 2D-CNN as input to a spatio-temporal Transformer (STT). Wang \textit{et al.}~\cite{wang2022dpcnet} proposed the dual path multi-excitation collaborative network (DPCNet). DPCNet consists of two modules, a spatial-frame excitation module to extract spatial features and a channel-temporal aggregation module to aggregate channel and temporal aware features.

\subsection{Ambiguity in FER}\label{sec:rw-amb}
Facial expressions are known to contain multiple emotion classes~\cite{zhouEmotionDistributionRecognition2015,dantcheva2017expression}. Ambiguous data are often regarded as noisy or inconsistent data. There are several kinds of approaches to deal with ambiguous data such as uncertainty estimation. She \textit{et al.}~\cite{sheDiveAmbiguityLatent2021} proposed an architecture with a latent label distribution module and uncertainty estimation module to address the ambiguity. Wang \textit{et al.}\cite{wang2020suppressing} proposed an extra module to suppress harmful instances and find latent truths by ordering with an estimated confidence level. 
Li \textit{et al.}~\cite{li2023intensity} proposed a global convolution attention block and intensity aware loss (GCA+IAL). 
IAL is designed to have the network pay extra attention to the most confusing category.
While their approach focuses on the uncertainty of facial expressions through attention blocks and loss functions, our approach handles ambiguity in DFER using data augmentation with soft labels. Soft labels are a simple approach against ambiguous data. Barsoum \textit{et al.}~\cite{zhouEmotionDistributionRecognition2015} investigated whether soft labels annotated by multiple crowd workers improved the static FER performance of a deep learning architecture and showed a model trained with soft labels outperformed that trained with hard labels. In addition, Gan \textit{et al.}~\cite{ganFacialExpressionRecognition2019} proposed a framework to generate pseudo soft labels for static FER. However, to the best of our knowledge, no prior research has focused on using a mixing strategy with soft labels for DFER.

\subsection{Mixing strategy}\label{sec:rw-mix}
The application of mixing strategies in data augmentation has widely been investigated. Zhang \textit{et al.}~\cite{zhang2018mixup} proposed a data augmentation method called \textit{mixup}, where additional training samples are synthesized by convexly combining random pairs of images and their labels. Mixup is based on the vicinal risk minimization~\cite{chapelle2000vicinal} principle, where the vicinity of each training sample is used to approximate the true distribution, thereby improving the generalization capability. Thulasidasan \textit{et al.}~\cite{thulasidasan2019mixup} showed that mixup is also a better training strategy from the viewpoint of confidence calibration. 

Inspired by mixup, other methods using mixing strategies have also been proposed. CutMix~\cite{yun2019cutmix} and CutOut~\cite{devries2017improved} use regional crop-and-paste techniques. Saliency information was employed in SaliencyMix~\cite{uddin2020saliencymix} and PuzzleMix~\cite{kim2020puzzle}. Some methods, such as Attentive-CutMix~\cite{walawalkar2020attentive} and TransMix~\cite{chen2022transmix}, employ activation or attention maps to achieve mixing. The mixing strategy is also applied to feature space in certain methods such as Manifold-mix~\cite{verma2019manifold} and PatchUp~\cite{faramarzi2020patchup}. 
MixGen~\cite{hao2023mixgen} enhances vision-language representation learning, employing multi-modal data augmentation based on image interpolation and text concatenation. Unlike our focus on label-based classification, MixGen targets tasks such as visual grounding and reasoning.

Most studies focus on image mixing, while few apply the mixing strategy to video data. Sahoo \textit{et al.}~\cite{sahoo2021contrast} proposed background mixing for contrastive learning in action recognition; however, their method was not used for generating data belonging to a class different from that of the source data. In addition, existing studies use hard labels because class information is clearly different from others in other computer vision tasks such as object and action recognition. However, at the time of writing, there has been no prior research dedicated to a mixing strategy for DFER and soft labels.

%% file: src/3_Method.tex
\begin{figure*}[t]
  \centering
  \includegraphics[width=0.72\linewidth]{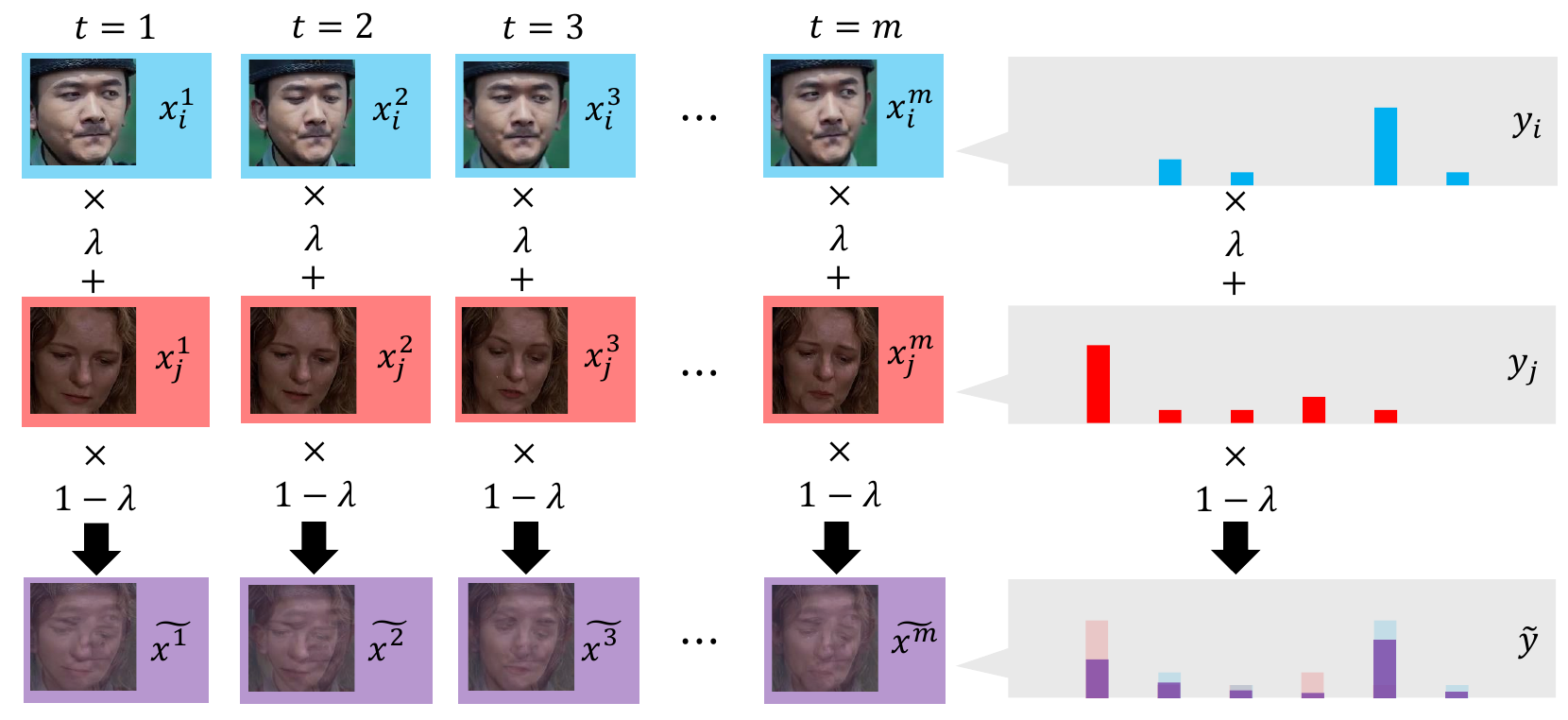}
  \vspace{0pt}
  \caption{Outline of the data mixing procedure in MIDAS. In MIDAS, the training data are augmented by convexly combining pairs of video frames and their corresponding emotion class labels. The mixing coefficient $\lambda$ is randomly generated from a beta distribution. The key point is that soft labels representing class probabilities are used instead of hard labels.}
  \vspace{-10pt}
  \label{fig:mixup}
\end{figure*}

MIDAS generates data in a similar way to mixup, that is, it convexly combines given training data and labels using a randomly generated mixing coefficient. It differs from mixup in that (i) the input data are video clips and (ii) soft labels representing class probabilities are given instead of single ground-truth class labels encoded in one-hot format, i.e., hard labels. The soft labels are assumed to be given based on the average of the votes by multiple annotators. This is due to the fact that recognizing ambiguous facial expressions is difficult even for the human eye, and each annotator's judgment is not necessarily correct. 

The important point is that the true hard label is unknown, and the proposed method is designed to minimize the vicinal risk under this condition. The data mixing procedure and how it minimizes vicinal risk are described below.

\subsection{Data mixing}\label{sec:datamixing}
Let $X_{i}=\left(x^{(1)}_i,\ldots,x^{(T)}_i\right)$ be the $i$-th video clip with a length of $T$ in the training dataset, where $x^{(t)}_i \in \mathbb{R}^{H\times W \times 3}$ is an image of height $H$ and width $W$ from the $t$-th frame in the video. 
In addition, let $y_i \in \mathbb{R}^C$ denote the soft-labeled ground truth for the $i$-th video clip whose elements are the probabilities for $C$ emotion classes.
MIDAS combines each frame of two video clips randomly selected from the training dataset and generates virtual samples, as illustrated in Fig.~\ref{fig:mixup}. 
The generated video clip $\Tilde{X}$ and label $\Tilde{y}$ are formulated as
\begin{flalign}
\Tilde{X}&=\left(\Tilde{x}^{(1)},\ldots,\Tilde{x}^{(T)}\right),\\
\Tilde{x}^{(t)}&=\lambda x^{(t)}_i + (1-\lambda)x^{(t)}_j,\\
\Tilde{y}&=\lambda y_i + (1-\lambda)y_j,
\end{flalign}
where $\lambda \in [0,1]\sim \beta (\alpha,\alpha)$ is a random ratio that follows the beta distribution with $\alpha$. It should be noted that MIDAS does not allow the same video to be selected. After that, to take annotation noise such as misjudgment into account~\cite{uma2020case}, a combined soft label $\Tilde{y}$ is normalized by using a softmax operation. By applying our method to the facial expression data, data that have multiple emotion classes with different intensities and temporal changes can be generated.

\subsection{Vicinal risk minimization}
The data augmentation that MIDAS performs is justified from the viewpoint of vicinal risk minimization~\cite{chapelle2000vicinal}, as is done in mixup~\cite{zhang2018mixup}. The difference with mixup is that the true hard label of each sample in the training data is unknown, and a soft label that includes the variation of the annotators' evaluation is used instead. Accordingly, it can be explained that MIDAS calculates an empirical risk using a distribution different from that of mixup.

In supervised learning, we assume a joint probability distribution $P(x,y)$ over input and output variables and minimize the expectation of a given loss function $\ell$.
\begin{equation}
\label{eq:riskMimization}
    R(f) = \int\ell(f(x),y)\mathrm{d}P(x,y),
\end{equation}
where $f$ is a classifier to be trained. However, eq.~(\ref{eq:riskMimization}) cannot directly be computed because the joint distribution $P(x,y)$ is unknown. In general, we minimize instead the empirical risk given a training dataset $\{\left(x_i,y_i\right)\}^M_{i=1} \sim P(x,y)$.
\begin{equation}
\label{eq:empiricalRisk}
    R_\mathrm{emp}(f) = \frac{1}{M}\sum^M_{i=1}\ell(f(x_i),y_i)
\end{equation}
The empirical risk is derived by taking an expectation of the loss function over an empirical distribution $P_\mathrm{emp}(x,y) = \frac{1}{M}\sum^M_{i=1}\delta(x=x_i,y=y_i)$, where $\delta$ is a Dirac measure. 

There are other possible choices to approximate the true distribution, and a different choice of distribution results in different risk minimization. The empirical vicinal risk based on the vicinity distribution~\cite{chapelle2000vicinal} is one of them. In~\cite{zhang2018mixup}, it was shown that mixup training minimizes the empirical vicinal risk:
\begin{equation}
    R_\mathrm{mixup}(f) = \frac{1}{M}\sum^M_{i=1}\ell(f(\tilde{x}_i),\tilde{y}_i),
\end{equation}
where $\{(\tilde{x}_i,\tilde{y}_i)\}^M_{i=1}$ is a set of virtual feature-target pairs generated from the vicinity distribution defined as
\small
\begin{flalign}
\label{eq:VicinityDistribution}
    &P_{\mathrm{mixup}}(\tilde{x}_i,\tilde{y}_i \mid x_i,y_i) \nonumber \\
    &\!=\! \frac{1}{n}\sum^n_j\underset{\lambda}{\mathbb{E}}[\delta(\tilde{x}_i \!=\! \lambda x_i \!+\! (1-\lambda)x_j, \tilde{y}_i \!=\! \lambda y_i \!+\! (1-\lambda)y_j)].
\end{flalign}
\normalsize

In the problem setting of this study, the hard label $y_i$\footnote{The superscript for the frame number is omitted for simplicity in this subsection.} corresponding to the underlying true emotion is unknown, and instead a soft label $q_i$ based on the voting average of multiple annotators is given. In MIDAS, the training data are sampled from the following distribution:
\small
\begin{flalign}
\label{eq:softVicinityDistribution}
    &P_{\mathrm{MIDAS}}(\tilde{x}_i,\tilde{y}_i \mid x_i,q_i) \nonumber \\
    &\!=\! \frac{1}{n}\sum^n_j\underset{\lambda}{\mathbb{E}}[\delta(\tilde{x}_i \!=\! \lambda x_i \!+\! (1-\lambda)x_j, \tilde{y}_i \!=\! \lambda q_i \!+\! (1-\lambda)q_j)].
\end{flalign}
\normalsize

Equation~(\ref{eq:softVicinityDistribution}) can be regarded as a variation of vicinity distribution. Assuming that $S$ annotators give one-hot labels $v_i^{(s)}$ ($s = 1, \ldots, S$) to each training sample $x_i$, the soft label $q_i$ is given by the average of the annotators' votes as $q_i = \frac{1}{S}\sum^S_{s=1}v_i^{(s)}$. If $l$ out of $S$ votes are correct, $q_i$ is expressed as $q_i = \frac{l}{S}y_i + \frac{1}{S}\sum_{s \in \mathcal{W}}v_i^{(s)}$, where $\mathcal{W}$ is a set of indices for wrong annotations. Using this expression, Eq.~(\ref{eq:softVicinityDistribution}) can be written as 
\small
\begin{flalign}
\label{eq:newSoftVicinityDistribution}
    &P_{\mathrm{MIDAS}}(\tilde{x}_i,\tilde{y}_i \mid x_i,q_i) \nonumber \\
    &\!=\! \frac{1}{n}\!\sum^n_j\underset{\lambda,\lambda'}{\mathbb{E}}\![\delta(\tilde{x}_i \!\!=\!\! \lambda x_i \!+\! (1\!-\!\lambda)x_j, 
    \tilde{y}_i \!\!=\!\! \lambda' y_i \!+\! (1\!-\!\lambda')y'_j)],
\end{flalign}
\normalsize
where we defined $\lambda' = \frac{\lambda l}{S}$ and $y'_j = \frac{\lambda}{S-\lambda l}\sum_{s \in \mathcal{W}}v_i^{(s)} + \frac{S(1-\lambda)}{S-\lambda l}q_j$. 
MIDAS corresponds to minimizing the vicinal risk in a situation where the true hard label $y_i$ is unknown, by defining the vicinity distribution using a random ratio and virtual labels that are different from the original mixup.

%% file: src/4_Experiments.tex
\label{sec:experiments}
The purpose of this experiment is to verify the validity of MIDAS for DFER using a deep learning-based automatic DFER model. 
We evaluated the performance of MIDAS using a publicly available DFER dataset, and compared the results with those of existing methods for DFER including the state-of-the-art one. 

\subsection{Evaluation dataset}	
We used the dynamic facial expression in-the-wild (DFEW) dataset, a collection of 11,967 video clips sourced from movies. The DFEW stands as the singular dataset providing soft labels for each individual video clip. These video clips contain various challenging interferences in practical scenarios such as extreme illumination, occlusions, and capricious pose changes. Twelve expert annotators were hired for this dataset, and ten out of twelve annotators were assigned to each video clip. Each annotator was asked to select one out of seven emotion classes (``happy,'' ``sad,'' ``neutral,'' ``angry,'' ``surprise,'' ``disgust,'' and ``fear''). The voting results by the annotators are provided as seven-dimensional emotion distribution labels, which are used as soft labels in this experiment. This dataset also stores the class with the highest number of votes by the annotators for each sample, and we used them as a hard label in comparative experiments.\par

\begin{figure}[t]
  \centering
  \includegraphics[width=0.8\linewidth]{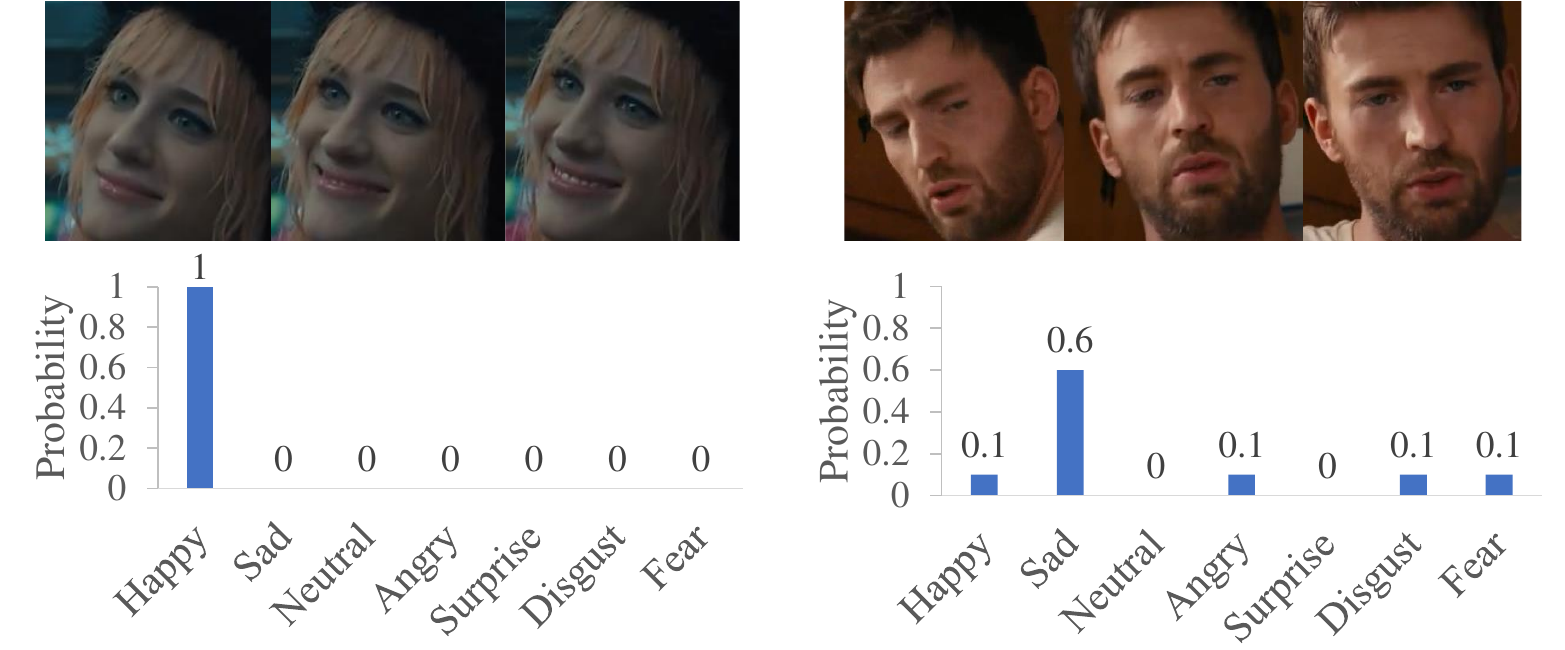}
  \vspace{-3pt}
  \caption{Examples of a clear facial expression (left) and ambiguous facial expression (right) with their soft label annotations in the DFEW dataset~\protect\cite{jiang2020dfew}}
  \vspace{-7pt}
  \label{fig:compare}
\end{figure}

Fig.~\ref{fig:compare} shows examples of facial expression images and the corresponding emotion labels in the DFEW dataset. In the figure, the left and right panels show examples of a clear expression that all annotators judged as ``happy'' and an ambiguous facial expression, respectively. In the example of an ambiguous expression on the right panel, the votes by the annotators are split into five classes although more than half of the annotators judged this sample as ``sad.''\par

\begin{figure}[t]
  \centering
  \includegraphics[width=0.75\linewidth]{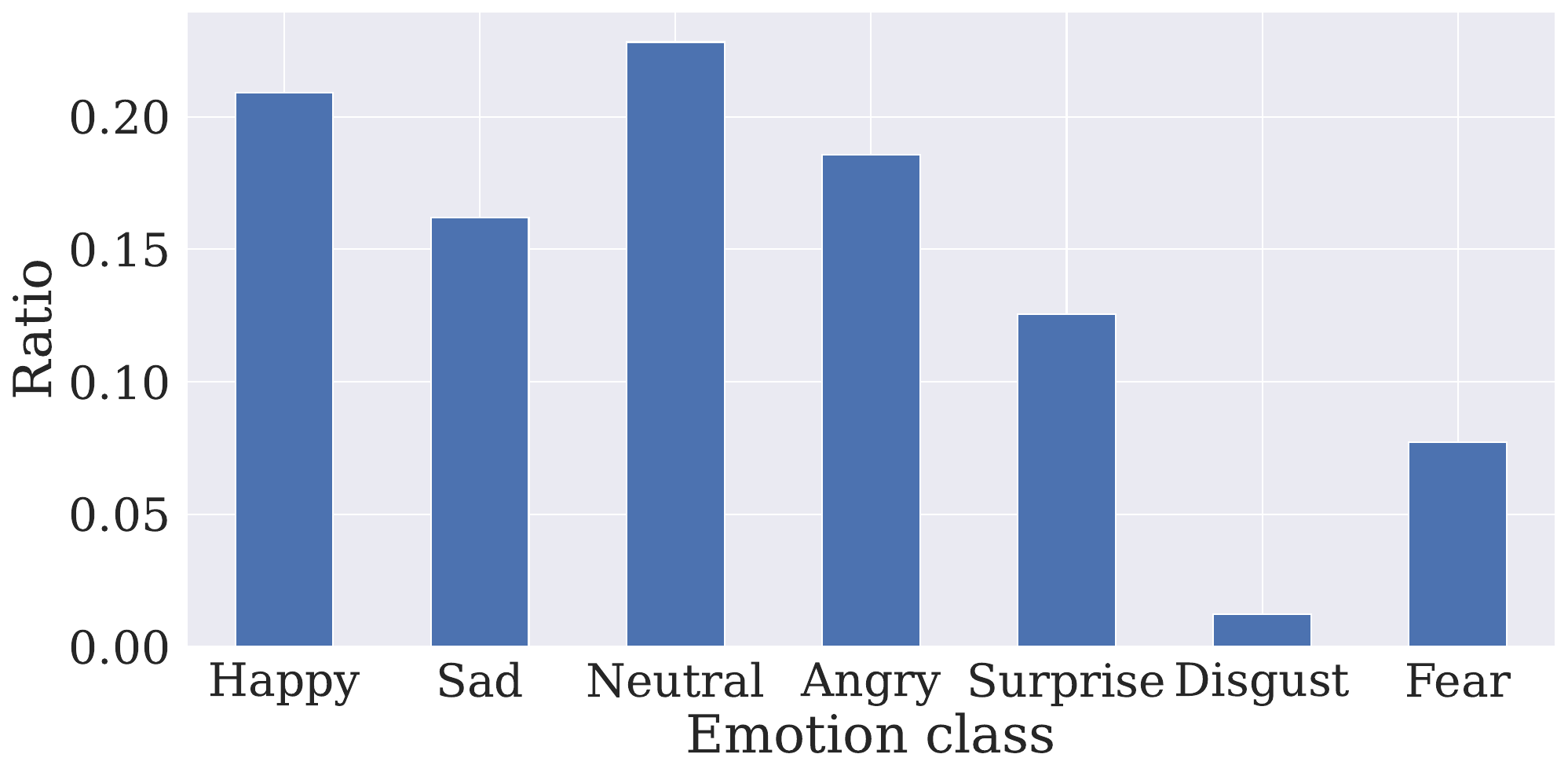}
  \vspace{-3pt}
  \caption{Emotion class distribution of the DFEW dataset}
  \vspace{-7pt}
  \label{fig:class}
\end{figure}

Fig.~\ref{fig:class} shows the distribution of emotion classes. The DFEW dataset is a class-imbalanced dataset that contains relatively more ``natural'' and ``happy'' and less ``disgust'' and ``fear.''

\subsection{Preprocessing}
First, we detected a facial region using Face++~\cite{facepp}. Face++ is a face recognition-related software that can be used for face detection, face comparison, and face retrieval, and we used its face detection function in this experiment. We then extracted facial landmarks from the detected face area and applied an affine transformation to landmarks to align the position of facial landmarks using Seeta~\cite{liu2017viplfacenet} by referring to the method in~\cite{jiang2020dfew}.

\subsection{Experimental conditions}\label{sec:exp-condition}
We used the temporal shifted module (TSM) with the ResNet-18 backbone~\cite{lin2019tsm}, which has been used for facial emotion recognition from video clips in previous studies. The ResNet-18 was pre-trained on ImageNet~\cite{russakovsky2015imagenet}. Since the length of the videos varied, we divided each video into eight segments and sampled one frame from each segment. The hyper-parameter settings for training were fixed. Regarding optimizers, we used SGD~\cite{ruder2016overview} with an initial learning rate of 0.02 and a momentum value of 0.9. Cosine learning rate decay was applied in training, and the number of training epochs was set to 450. We applied random scaling before inputting the video clip into the model at training time. We used a batch size of eight and a dropout with a rate of 0.5. The input image of each frame was resized to 224 $\times$ 224. For the loss function, we used cross-entropy loss. In addition, $\alpha$ for the beta distribution for MIDAS was set to 0.4.\par

To evaluate the generalization capability, we performed five-fold cross-validation using the data split provided in the DFEW dataset. For the evaluation indexes, we employed the unweighted average recall (UAR) and weighted average recall (WAR), which are officially used in~\cite{jiang2020dfew}. UAR represents the average prediction accuracy of each class and WAR represents accuracy. We calculated the averages of UAR and WAR over five groups of cross-validation. We compared the results with those of some existing methods for DFER proposed in~\cite{hara2018can,zhao2021former,he2016deep,chung2014empirical,ma2022spatio} including the state-of-the-art method. We also employed the ResNet-18 with TSM trained simply on soft and hard labels for comparison methods to evaluate the effectiveness of MIDAS.

\subsection{Result}\label{ambiguity}\label{sec:ex-res}
Table~\ref{tab:each_class} summarizes the accuracy for each emotion class, the WAR, and UAR, with the scores of comparative methods and our model trained on the original video (i.e., without data augmentation) with soft and hard labels. MIDAS achieved the best scores in WAR and UAR, thereby showing the effectiveness of the combination of soft labels and mixing strategy for DFER.
\begin{table*}[t]
\centering
    \caption{Comparison of the accuracy for each emotion class, UAR, and WAR. Bold and underlined scores denote the best and second best, respectively.}
    \vspace{-3pt}
    \label{tab:each_class}
    \resizebox{\textwidth}{!}{%
    \scalebox{1.0}[1.0]{
    \begin{tabular}{p{0.3\linewidth}|c|ccccccc|c|c} 
    \toprule
         \multirow{2}{*}{Method} &\multirow{2}{*}{Label} &  \multicolumn{7}{c}{Accuracy for each emotion class (\%)} & \multicolumn{2}{|c}{Metrics} \\ \cmidrule(lr){3-9}\cmidrule(lr){10-11}
             & & happy & sad & neutral & angry & surprise & disgust & fear & UAR & WAR\\
     \midrule  
          \rowcolor[rgb]{0.9, 0.9, 0.9}
         3D Resnet18~\cite{hara2018can,zhao2021former}& Hard  & 79.18 & 49.05 & 57.85 & 60.98 & 46.15 & 2.76 & 21.51 & 46.52 & 58.27 \\
         
        Resnet18+GRU~\cite{he2016deep,chung2014empirical,zhao2021former}
        & Hard  & 82.87 & 63.83 & 65.06 & 68.51 & 52.00 & 0.86 & 30.14 & 51.68 & 64.02\\
         \rowcolor[rgb]{0.9, 0.9, 0.9}
         Former-DFER~\cite{zhao2021former} & Hard  &  84.05 & 62.57 & \underline{67.52} & 70.03 & 56.43 & 3.45 & 31.78 & 53.69 & 65.70\\
         
         STT~\cite{ma2022spatio} & Hard  &  87.36 & \bf{67.96} & 64.97 & \underline{71.24} & 53.10 & 3.49 & \underline{34.04} &54.58& 66.65 \\
         \rowcolor[rgb]{0.9, 0.9, 0.9}
         DPCNet~\cite{wang2022dpcnet} & Hard  &  \bf{89.59} &  64.82 & 66.98 & 63.14 & 53.81 & \underline{14.48} & 32.34 & \underline{57.11} & 66.32\\
         GCA+IAL~\cite{li2023intensity} & Hard & \underline{87.95} & 67.21 & \bf{70.10} & \bf{76.06} & \bf{62.22} & 0.00 & 26.44 & 55.71 & \bf{69.24} \\
         
         \rowcolor[rgb]{0.9, 0.9, 0.9}
         Resnet18+TSM (Ours)& MIDAS &  87.40 & \underline{67.34 }& 58.64 & 68.06 &\underline{59.65}& \bf{28.69} &\bf{44.50} & \bf{57.45} & \underline{69.16}\\
    \bottomrule
    \end{tabular}
    }}
\end{table*}

Compared to other methods, MIDAS showed the best score in UAR with a 0.34\% gap from the second-best method (DPCNet) and achieved the second-best score in WAR with only a 0.08\% gap from GCA+IAL, which is the state-of-the-art method for the DFEW dataset. These results demonstrated the effectiveness of MIDAS in improving the performance of DFER.\par

Regarding the accuracy for each class, It should be emphasized that MIDAS scored higher than the other methods in ``disgust'' and ``fear,'' whose number of training samples is much less than that of the other classes (see Fig.~\ref{fig:class}). In contrast, GCA+IAL~\cite{li2023intensity}, which represents the state-of-the-art in WAR and is another approach addressing facial expression ambiguity based on the intensity-aware loss function, showed remarkably low accuracy in these categories. These results indicate that our method improves the accuracy for emotion classes with smaller data sizes.

\begin{figure*}[t]
  \centering
  \includegraphics[width=0.68\linewidth]{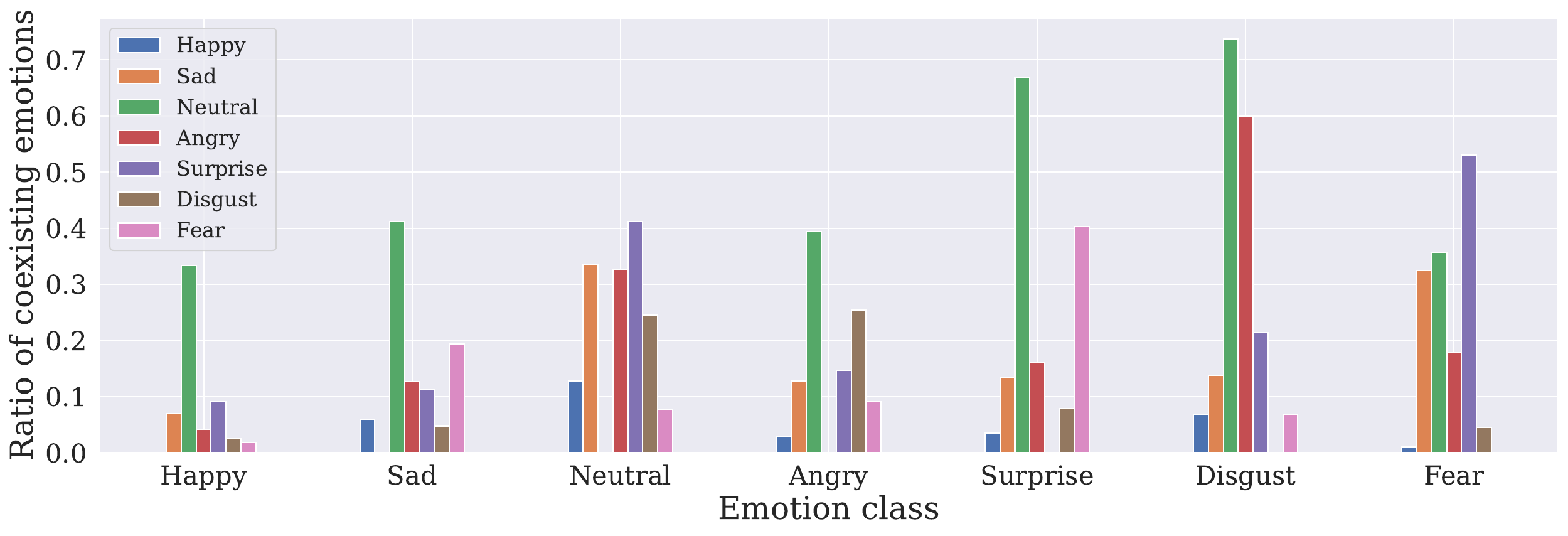}
  \vspace{-3pt}
  \caption{Ratio of coexisting emotions for each emotion class. The values in the figure were calculated by averaging the soft label values of the samples that belong to the corresponding emotion class. The higher the value, the more likely the emotion class is to be voted for by the annotators simultaneously, that is, the more likely it is to coexist.}
  \vspace{-5pt}
  \label{fig:dist}
\end{figure*}

%% file: src/5_AblationStudy.tex
\label{sec:ablation}
\subsection{Comparison with the model trained on soft and hard labels}\label{sec:ab-sh}
\vspace{-2mm}
While our approach has yielded noteworthy outcomes, it is essential to clarify MIDAS's effectiveness compared to models trained solely on hard and soft labels. To evaluate the effectiveness of MIDAS, we employed the ResNet-18 with TSM trained simply on soft and hard labels. The results are shown in Table~\ref{tab:ab-soft-hard}. Regarding WAR, MIDAS achieved 69.16\%, which outperforms the score of the models trained solely on hard labels (64.31\%) and soft labels (67.27\%). For UAR, MIDAS (57.45\%) marked a better score than the models with soft labels (54.61\%) and hard labels (54.03\%). These results demonstrate the effectiveness of MIDAS in improving the performance of DFER. In addition, compared with the existing methods in Table~\ref{tab:each_class}, our model with soft labels achieved the third-best in both WAR and UAR, suggesting the effectiveness of simply using soft labels even without data augmentation.
\begin{table}[t]
\caption{Comparison of the results of our method with the model trained on soft and hard label}
\centering
    \label{tab:ab-soft-hard}
    \vspace{-1mm}
    \scalebox{0.9}[0.9]{
    \begin{tabular}{c|c|c} 
    \toprule
         Label & UAR & WAR \\ 
        \midrule
            Hard   & 54.03 & 64.31\\
            Soft   & 54.61 & 67.27\\
            \begin{tabular}{c}
             MIDAS
            \end{tabular} & \textbf{57.45} & \textbf{69.16}\\
    \bottomrule
    \end{tabular}
    }
   \vspace{0pt}
\end{table}

\subsection{Our method with hard labels}\label{sec:ab-hard}
We investigated whether a mixing strategy for dynamic facial expressions with hard labels could improve performance. Creating soft labels is costly although our method can improve performance. It would be helpful if training on the single-labeled annotation that can be obtained more easily than soft labels also improves performance. 

The procedure of this experiment is simple; we applied our method to video clips with hard labels instead of soft ones. The model was trained on this generated dataset after normalizing the combined labels in the same settings described above. 

The results are shown in Table~\ref{tab:ab-midas-with-hard}. The model trained on MIDAS with hard labels improved the WAR of the model trained on the original dataset with hard labels by 1.35\%. For UAR, the score is 54.93\%, which is higher than a model trained on the original dataset with hard and soft labels. This result indicates that our strategy can enhance the performance of FER when using hard labels.

\begin{table}[t]
\caption{Comparison of the results of our method with and without hard label}
\centering
    \label{tab:ab-midas-with-hard}
    \vspace{-1mm}
    \scalebox{0.9}[0.9]{
    \begin{tabular}{c|c|c} 
    \toprule
         Label & UAR & WAR \\ 
        \midrule
            Hard   & 54.03 & 64.31\\
            Soft   & 54.61 & 67.27\\
            \begin{tabular}{c}
             MIDAS w/ hard label
            \end{tabular} & 54.93 & 65.66\\
            \begin{tabular}{c}
             MIDAS
            \end{tabular} & \textbf{57.45} & \textbf{69.16}\\
    \bottomrule
    \end{tabular}
    }
   \vspace{-7pt}
\end{table}
Furthermore, we extended our investigation to another large-scale dataset, FERV39k~\cite{wang2022ferv39k}. Unlike the DFEW dataset, only hard-labeled annotations are given in FERV39k. The experimental setup remains consistent with Section~\ref{sec:exp-condition}, except for the initial learning rate, set at 0.01. Table~\ref{tab:ab-ferv} demonstrates that MIDAS with hard labels achieved a superior UAR score (39.2\%) compared to current state-of-the-art methods on the FERV39k dataset, and secured the second-best score (47.37\%) in WAR. These results suggest that MIDAS possesses the potential for generalization even in scenarios involving only hard labels.
\begin{table}[t]
\caption{Comparison of the results of our method with hard label on FER39k~\cite{wang2022ferv39k}}
\centering
    \label{tab:ab-ferv}
    \vspace{-1mm}
    \scalebox{0.9}[0.9]{
    \begin{tabular}{l|c|c|c} 
    \toprule
         Method & Label & UAR & WAR \\ 
        \midrule
            Two VGG13-LSTM& Hard & 31.28&43.2 \\
            \rowcolor[rgb]{0.9, 0.9, 0.9}
            Former-DFER~\cite{zhao2021former} & Hard   & \underline{37.20} & 46.85\\
            GCA+ICL\cite{li2023intensity} & Hard & 35.82 & \bf{48.54}\\
            \rowcolor[rgb]{0.9, 0.9, 0.9}
            Resnet18+TSM (Ours) &\begin{tabular}{c}
             MIDAS \\ w/ hard label
            \end{tabular} & \bf{39.2} & \underline{47.37}\\
            
    \bottomrule
    \end{tabular}
    }
   \vspace{0pt}
\end{table}

\subsection{Cross-dataset evaluation}
Generalization ability is crucial in facial expression recognition due to the need to accommodate variations such as lighting conditions, facial orientations, and the diversity in facial shapes. These factors can significantly influence the accuracy of recognition models in real-world situations. 

We conducted a cross-dataset evaluation to examine whether our approach improves the model's generalization ability to unseen conditions. In this experiment, we trained the model on the DFEW dataset and tested it on the acted facial expressions in the wild (AFEW) dataset~\cite{dhallCollectingLargeRichly2012} to assess the performance on first-time encountered conditions. As in Section~\ref{sec:experiments}, we trained the TSM with ResNet-18 using MIDAS and compared its performance with those trained simply on soft and hard labels. 
These models were not further fine-tuned using the AFEW dataset, but simply evaluated on the AFEW test set.
Due to the absence of a method providing results for models trained on DFEW and tested on AFEW datasets, our comparison involves simply contrasting the outcomes of MIDAS using models trained with soft and hard labels.

Table~\ref{tab:afew} presents the UAR and WAR on the AFEW test set for each method. MIDAS demonstrated superior performance in both UAR and WAR compared to the models trained on hard and soft labels. These results suggested that MIDAS potentially improves the model's generalization ability to unseen conditions.

\begin{table}[t]
\caption{Comparison of our method's results on the AFEW test set}
\centering
    \label{tab:afew}
    \vspace{-1mm}
    \scalebox{0.9}[0.9]{
    \begin{tabular}{c|c|c} 
    \toprule
         Label & UAR & WAR \\ 
        \midrule
            Hard   & 38.20 & 40.72\\
            Soft   & 36.67 & 39.61\\
            \begin{tabular}{c}
             MIDAS
            \end{tabular} & \textbf{39.56} & \textbf{43.77} \\
    \bottomrule
    \end{tabular}
    }
   \vspace{-7pt}
\end{table}

\subsection{Analysis of the impact of coexisting emotion}
we analyzed the effect of coexisting emotions in DFER. The soft labels in the DFEW dataset were constructed based on the votes of ten annotators. During this voting process, the votes of all annotators do not necessarily coincide; some minority annotators vote for emotion classes other than the class that received the most votes. For example, ``sad'' received the most votes with eight votes, but ``angry'' also received two votes. We analyzed how such coexisting emotion classes that are often voted together affect the model capability.

Fig.~\ref{fig:dist} shows the average ratio of coexisting emotions for each emotion class. For example, in the case of the leftmost emotion class ``happy,'' ``neutral'' is also voted for a lot for the instances where ``happy'' received the most votes. The values in the figure were calculated by averaging the soft label values of the samples that belong to the corresponding emotion class. From this figure, the following are observed.
\begin{itemize}
\setlength{\parskip}{0cm}
\setlength{\itemsep}{0cm}
    \item ``Neutral'' tends to coexist with all other emotion classes.
    \item ``Happy'' does not coexist much with other emotion classes.
    \item ``Angry'' and ``disgust'' coexist frequently.
    \item ``Sad,'' ``surprise,'' and ``fear' often coexist.
\end{itemize}

\begin{figure}[t]
  \centering
  \includegraphics[width=0.78\linewidth]{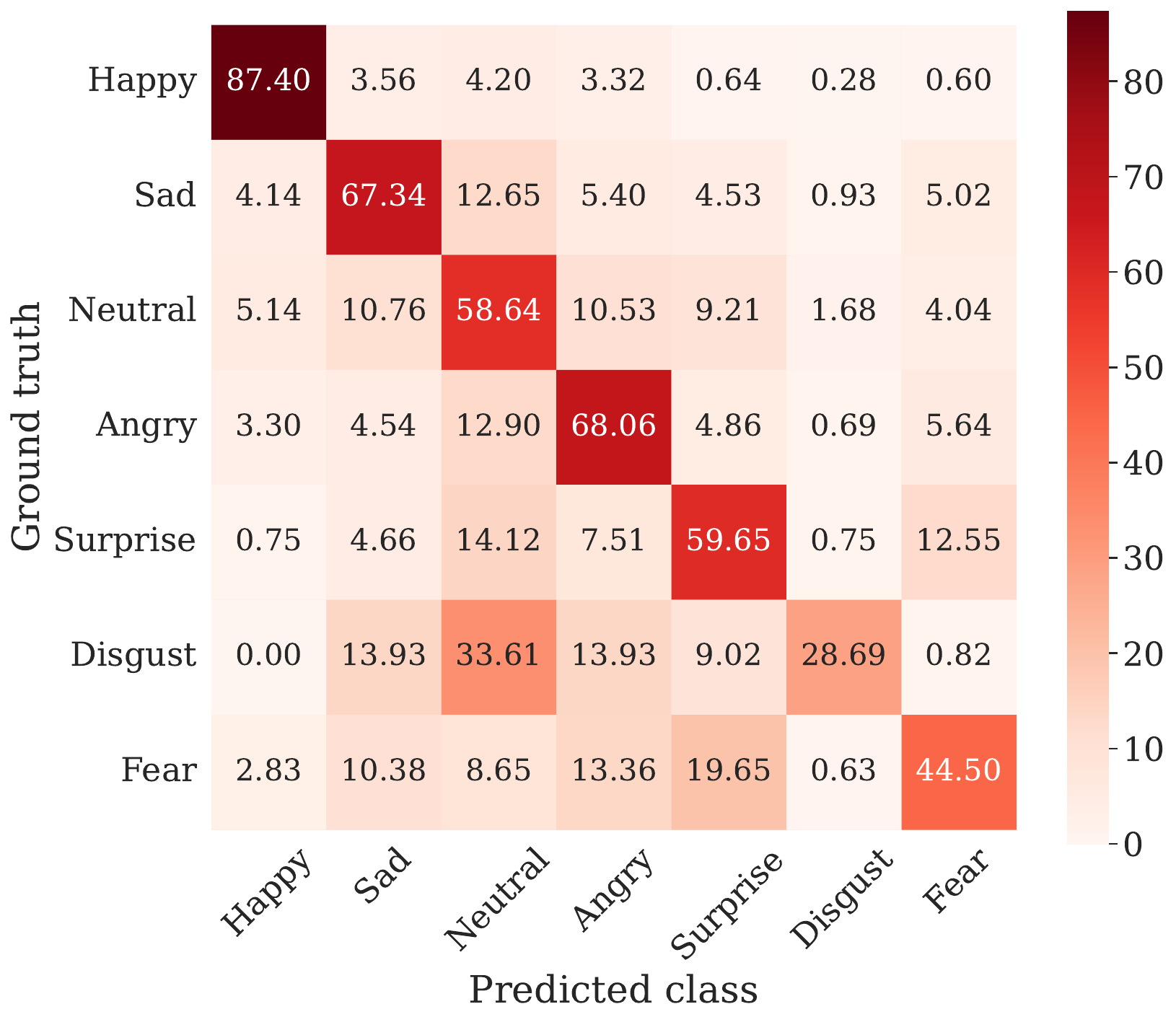}
  \vspace{0pt}
  \caption{Confusion matrix of the model trained with MIDAS}
  \vspace{0pt}
  \label{fig:conf}
\end{figure}

Fig.~\ref{fig:conf} shows the confusion matrix of the Resnet-18 with TSM model trained with MIDAS. The impact of emotion class coexistence revealed in Fig.~\ref{fig:dist} on the classification results can be observed. ``Neutral'' coexists with other classes more than other emotion classes, the model wrongly predicted many samples of other classes as neutral. In particular, instances classified as ``disgust'' are most often recognized as ``neutral.'' There are few coexisting emotions for ``happy,'' and the instances classified as ``happy'' are almost always predicted correctly (87.40\%). These results indicate that coexisting emotions affected model performance in the DFER.




\vspace{0pt}
\subsection{The effect of ambiguous data on model performance}
To confirm whether ambiguous data in the DFEW dataset affect the model performance, we investigated the effect of ambiguous data by comparing models trained on datasets with and without ambiguous data. In this comparison, we divided the original DFEW dataset into two groups: clear expression and mixed expression groups. The clear expression group consists of data with maximum soft label values of more than 0.9, e.g., the left example in Fig.~\ref{fig:compare}. The mixed expression group contains data regardless of the soft labels' values and includes ambiguous facial expressions such as the right example in Fig.~\ref{fig:compare}. To address the difference in data distribution, the distribution of each emotion class was matched to the original dataset by oversampling and down-sampling. Finally, the sizes of both data groups were set to an equal number (4275). We trained two models on these datasets with soft labels and evaluated them using a validation split of the original dataset. 

The results are shown in Table~\ref{tab:ab1}. The model with the mixed expression group obtained higher UAR and WAR scores than the model trained with the clear expression group. These results demonstrated that the existence of ambiguous data can improve the performance of DFER.
\begin{table}[!t]
\centering
    \caption{Comparison of the results of models trained with and without ambiguous data}
    \label{tab:ab1}
    \vspace{-1mm}
    \scalebox{0.9}[0.9]{
    \begin{tabular}{c|c|c} 
    \toprule
         Data & UAR & WAR \\ 
        \midrule
          Clear expression group & 45.42& 58.46 \\
         
          Mixed expression group & \textbf{48.54}& \textbf{60.91} \\
    \bottomrule
    \end{tabular}
    }
    \vspace{0pt}
\end{table}

\subsection{Comparison of different architectures}
\begin{table}[t]
\centering
    \caption{Results using the 2DCNN-GRU architecture, which is different from the architecture used in Table~\ref{tab:each_class}}
    \vspace{-1mm}
    \label{tab:ab3}
    \scalebox{0.9}[0.9]{
    \begin{tabular}{c|c|c} 
    \toprule
         Label & UAR & WAR \\ 
        \midrule
        Hard   & 51.62 & 64.00\\
        Soft   & 51.82 & 65.00\\
        MIDAS & \textbf{53.70} & \textbf{67.01}\\
    \bottomrule
    \end{tabular}
    }
    \vspace{-7pt}
\end{table}

In our experiments, we used TSM~\cite{lin2019tsm}, an architecture for dynamic data, and the results show our method with TSM improves the performance of DFER. 
However,  we did not investigate whether our method is effective for different deep learning architectures. 

We, hence, conducted an experiment with different architectures for DFER. 
In this experiment, we used a 2DCNN-GRU as a different deep learning architecture.
The 2DCNN-GRU was trained using MIDAS and compared with a model trained on the original dataset with soft and hard labels. The settings of training were the same as those in the other experiments. 

Table~\ref{tab:ab3} summarizes the results using the 2DCNN-GRU architecture. The scores of MIDAS with 2DCNN-GRU (UAR: 53.70\% and WAR: 67.01\%) are higher than those of the soft- and hard-label supervised models. In addition, its WAR is higher than those of the existing works except for GCA+IAL~\cite{li2023intensity} (see Table~\ref{tab:each_class}). These results indicate that MIDAS has the potential to improve the performance of DFER regardless of the architecture.

%% file: src/6_Conclusion.tex
\vspace{-5pt}
In this paper, to handle various ambiguous facial expressions, we proposed a data augmentation method called MIDAS, which is based on data mixing with soft labels for DFER. In our method, we combine two video clips of facial expressions and their soft labels to generate various combinations of emotions and intensities. We conducted experiments to evaluate our method with a dataset for DFER. The results showed that our method can enhance the performance of DFER and outperform the state-of-the-art method. 

In future work, we plan to evaluate MIDAS using other domains of datasets, as it was evaluated using only the DFEW dataset. Although the MIDAS is developed for DFER with ambiguous facial expressions, it would be effective for other tasks that involve ambiguous class categorization soft-labeled annotations. 